# An Improved Lightweight YOLOv5 Model Based on Attention Mechanism for Face Mask Detection

Sheng Xu[1], Zhanyu Guo[2], Yuchi Liu[3], Jingwei Fan[2], Xuxu Liu[1]

[1] Northwestern Polytechnical University, Xi'an, China
`xus@mail.nwpu.edu.cn`
[2] University of Leeds, Leeds, UK
[3] SiChuan University, Chengdu, China

**Abstract.** Coronavirus 2019 has brought severe challenges to social stability and public health worldwide. One effective way of curbing the epidemic is to require people to wear masks in public places and monitor mask-wearing states by utilizing suitable automatic detectors. However, existing deep learning based models struggle to simultaneously achieve the requirements of both high precision and real-time performance. To solve this problem, we propose an improved lightweight face mask detector based on YOLOv5, which can achieve an excellent balance of precision and speed. Firstly, a novel backbone ShuffleCANet that combines ShuffleNetV2 network with Coordinate Attention mechanism is proposed as the backbone. Afterward, an efficient path aggression network BiFPN is applied as the feature fusion neck. Furthermore, the localization loss is replaced with α-CIoU in model training phase to obtain higher-quality anchors. Some valuable strategies such as data augmentation, adaptive image scaling, and anchor cluster operation are also utilized. Experimental results on AIZOO face mask dataset show the superiority of the proposed model. Compared with the original YOLOv5, the proposed model increases the inference speed by 28.3% while still improving the precision by 0.58%. It achieves the best mean average precision of 95.2% compared with other seven existing models, which is 4.4% higher than the baseline.

**Keywords:** Face mask detection, YOLOv5, Attention mechanism, Feature extraction and fusion

## 1 Introduction

Since December 2019, the coronavirus disease 2019 (COVID-19) epidemic has spread rapidly around the whole world, posing a great threat to public health. Wearing face masks in public places can help prevent the spread of the virus; nowadays, governments have mandated people to wear masks when going out. To supervise mask-wearing in crowded public areas, most places still adopt traditional measures such as manual inspection, which wastes human resources and may cause lapses in inspection. To address the existing problems encountered in traditional measures, face mask detection models based on deep learning have attracted extensive attention from researchers.

Some algorithms have achieved relatively good performance. Shylaja et al. [1] used transfer learning technique to train their face mask detection model based on Faster R-CNN, which exhibited excellent identification performance on images of indoor environments and complex scenes. However, it could hardly realize real-time detection and was difficult to be deployed on portable or mobile devices. Fan et al. [2] proposed RetinaFaceMask, a one-stage face mask detector with MobileNet as the backbone. It could distinguish between correct and incorrect mask-wearing states, but still did not solve the problem of precision reduction caused by the light weight.

In this paper, we propose a high-performance, lightweight face mask detector based on YOLOv5 [3] and attention mechanism to detect whether people wear masks. The overview working pipeline of the proposed model is shown in Fig. 1. We use ShuffleNetV2 [4] as the backbone and increase the kernel size from three to five in each depthwise convolution for a larger receptive field. Furthermore, we integrate Coordinate Attention [5] into it, forming a novel backbone called ShuffleCANet that is still lightweight but more effective with attention mechanism. Then we replace the original PANet with BiFPN [6] as the feature fusion neck to fully use multi-scale features. Moreover, we replace the localization loss with α-CIoU [7], which is more robust to lightweight models. Additionally, we also utilize several valuable methods for data processing. Generally speaking, the proposed model can achieve a balance of high accuracy and fast speed, providing an excellent solution to handle the problems associated with heavy models (difficult to deploy) and light models (struggle to achieve satisfactory performance). Thus it is a more practical alternative for deployment on relevant devices, thereby contributing to the battle against COVID-19. The main contributions of this paper are listed as follows:

1) We propose a novel backbone called ShuffleCANet with Coordinate Attention mechanism, which is lightweight but has a strong feature extraction ability. Moreover, we use BiFPN as the feature fusion neck to make full use of the features between different scales.
2) We modify the localization loss with α-CIoU to make bounding boxes regress better and obtain higher quality anchors.
3) We evaluate the proposed model on the AIZOO dataset [8]. Experimental results demonstrate that the proposed model can achieve state-of-the-art performance with real-time speed and high precision, surpassing seven other existing models.

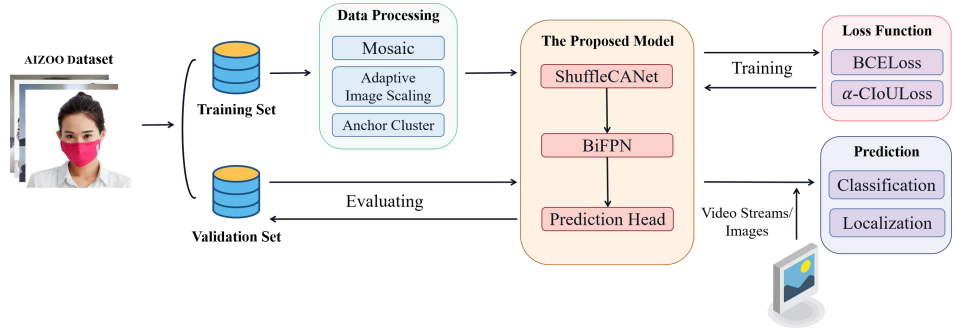

**Fig. 1.** The overview working pipeline of the proposed model.

The remainder of this paper is organized as follows. In Section 2, some related works about object detection and face mask detection are reviewed. In Section 3, the detailed architecture of the proposed methodology is presented. In Section 4, the implementation details of the proposed work are described. Finally, in Section 5, conclusions of the whole work are presented and future works are discussed.

## 2 Related Works

### 2.1 Object Detection

Traditional object detection algorithms are primarily based on manually designed features such as Haar [9] and Hog [10] features. However, these features have weak generalization ability and poor robustness. In recent years, deep learning based algorithms have become the mainstream research method for object detection because of their excellent performance. The development history of them is shown in Fig. 2.

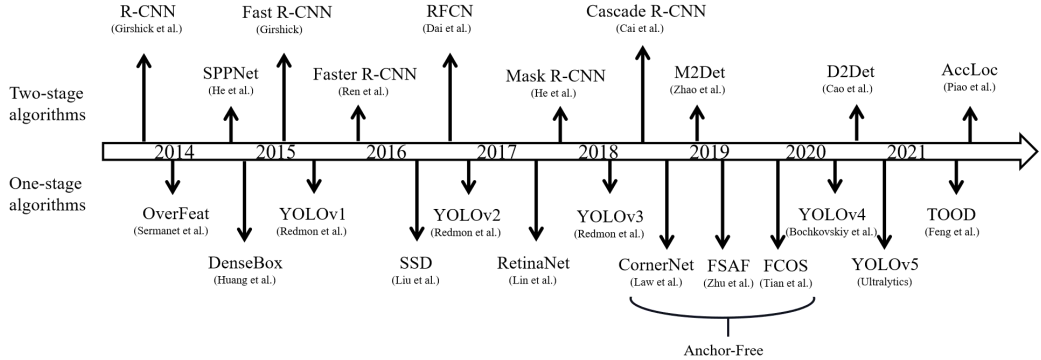

**Fig. 2.** The development history of deep learning based object detection models.

Deep learning based object detection algorithms can be mainly divided into two categories: two-stage and one-stage algorithms. Two-stage algorithms first generate a group of region proposals that may contain targets, and then further classify and calibrate them. They have high precision but poor real-time performance, represented by the R-CNN [11-13] series. One-stage detectors are based on the idea of regression, they directly predict the coordinates and categories of targets in a single step. They have fast speeds to meet real-time requirements but inferior precision, represented by the YOLO [14-17] series, SSD [18], RetinaNet [19], etc. Moreover, some novel anchor-free algorithms based on dense prediction have been prevalent in recent years, such as CornerNet [20] and FCOS [21]. They detect objects as paired keypoints or in a per-pixel prediction fashion instead of generating prior anchor boxes, thus avoiding the complicated computation related to anchors.

### 2.2 Face Mask Detection

Research on face mask detection has been prevalent since the eruption of COVID-19, because using face mask detectors to automatically monitor mask-wearing states in public places is an efficient way to help curb the epidemic.

As for two-stage detectors, Loey et al. [22] proposed a hybrid model comprising two components for face mask detection: the first for feature extraction and the second for classification, which performed well on three public face mask datasets. In [23], the authors proposed a dual-stage convolutional neural network architecture capable of detecting masked and unmasked faces, which could be pre-installed in CCTV cameras. As for one-stage detectors, YOLO-based models have been extensively studied for face mask detection. Several methods have been applied to YOLO series to increase their performance, as noted in [24-27]. Among them, attention mechanism is widely used because it can focus on useful information and enhance the prediction ability of the model, such as convolutional block attention module (CBAM) [28] and Squeeze-and-Excitation Block (SE) [29]. Besides, the work described in

[30] proposed a novel single-shot lightweight face mask detector (SL-FMDet) using MobileNet and an improved FPN with residual context attention module and Gaussian heatmap regression.

However, existing face mask detection algorithms pose some problems. On the one hand, some powerful detectors can yield extremely high precision but are too heavy and costly to be used in practice. On the other hand, some lightweight detectors are easy to deploy but struggle to achieve high-precision detection.

## 3 Methodology

### 3.1 Data Processing

To better train the model, utilizing some methods in the training phase to process the original data is significant. The data processing methods employed in this paper include Mosaic [17], adaptive image scaling, anchor cluster and some other relevant methods.

Firstly, Mosaic method is applied to the original AIZOO training set, splicing four images by random scaling, and then cropping and arranging them into a single image; this increases the diversity and enriches the background of the picture. Then the adaptive image scaling operation is applied to obtain the 640×640 standard image for training. Specifically, the longer side of the input image is resized to 640 pixels while maintaining the aspect ratio, and then the image is overlaid on a generated 640×640 grayscale image. Moreover, in case of the large difference in size between anchors and objects, k-means algorithm is used to cluster the bounding boxes of the training set. The automatically generated prior anchors are more suitable and efficient, which can learn more effective prior knowledge and accelerate model convergence.

### 3.2 Network Architecture

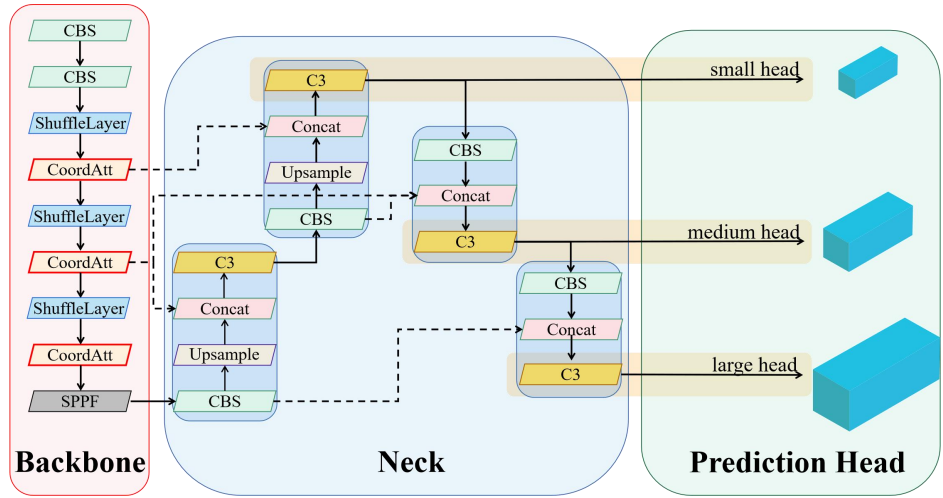

**Fig. 3.** The network architecture of the proposed model. a) The backbone is composed of modified ShuffleNetV2 and three CoordAttention modules. The CBS module consists of a Conv2d layer, a BatchNorm layer, and a SILU activation. The detailed structure of ShuffleLayer and CoordAttention are illustrated in Fig. 4 and Fig. 5. b) The neck use BiFPN structure to fuse features. c) The three prediction heads are used for small, medium, and large objects, respectively.

In this paper, the detection task is treated as a regression problem based on YOLOv5, and the model is further optimized for face mask detection. The network architecture of the model is shown in Fig. 3. The key improvements in architecture are summarized as follows: we propose a novel backbone called ShuffleCANet based on modified ShuffleNetV2 [4] and CoordAttention [5] mechanism, which is lightweight but still achieve SOTA performance; we utilize BiFPN [6] to replace PANet as the path aggression network for more efficient multi-scale feature fusion. To sum up, the model is lighter, faster, and more accurate, which is more efficient and practical to deploy.

**Modified ShuffleNetV2.** ShuffleNet [31] is a highly computation-efficient CNN architecture for mobile devices with point-wise group convolution and channel shuffle operations. ShuffleNetV2 is an improved version of ShuffleNet. It introduces channel split operation and changes the element-wise addition by concatenation, followed by the channel shuffle operation to ultimately mix features.

In this paper, we use ShuffleNetV2 as the basic backbone and implement several modifications to improve its performance. To be specific, we retain the original two CBS blocks with small parameters to ensure less down-sampling loss in the initial stage and guarantee the learning ability. Moreover, in each ShuffleLayer, the kernel size of depthwise convolution is increased from 3 to 5 for a larger receptive field. The structure of ShuffleLayer in this work is shown in Fig. 4.

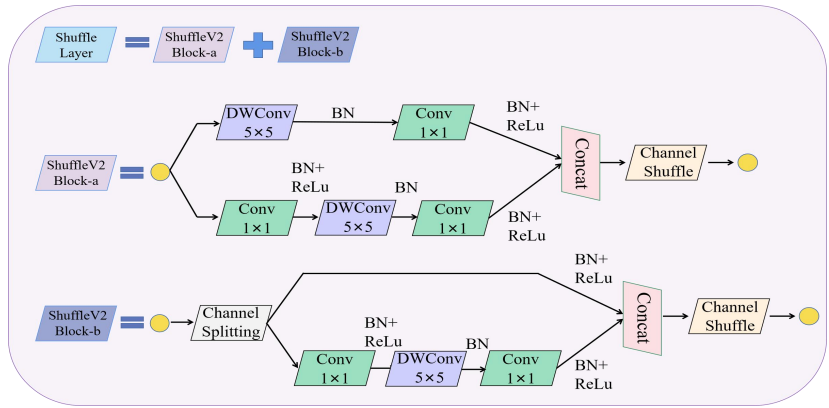

**Fig. 4.** The structure of ShuffleLayer. DWConv denotes depthwise convolution.

**CoordAttention.** CoordAttention [5] is a novel, lightweight, and efficient attention module of attention mechanism to enhance model performance, which can be easily incorporated into mobile networks to improve the accuracy with little additional computing overhead. The structure of CoordAttention module is shown in Fig. 5.

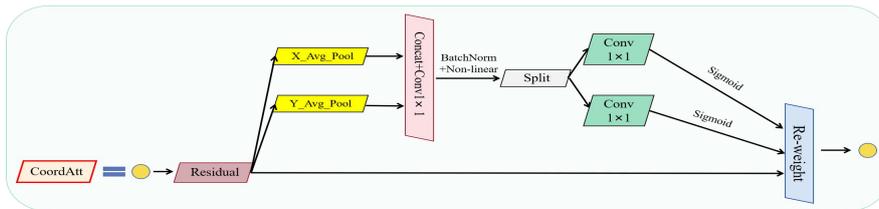

**Fig. 5.** The structure of CoordAttention. X_Avg_Pool and Y_Avg_Pool denote 1D horizontal global pooling and 1D vertical global pooling, respectively.

To be specific, the operation is divided into two steps: coordinate information embedding and coordinate attention generation. The first step factorizes channel attention into two 1D global pooling processes to encode each channel along the horizontal and vertical directions, respectively. The formulas are as follows:

$$Z_c^h(h) = \frac{1}{W}\sum_{0 \leq i < W} x_c(h, i) \ , \ Z_c^w(w) = \frac{1}{H}\sum_{0 \leq i < H} x_c(j, w) \tag{1}$$

where X denotes the input, $Z_c^h(h)$ and $Z_c^w(w)$ denote the output of the $c$-th channel at height $h$ and width $w$, respectively. In the second step, the produced feature maps are then concatenated and sent to a shared 1×1 convolutional transformation $F_1$ to obtain the intermediate feature map $f$ as follows:

$$f = \delta(F_1([Z^h, Z^w])) \tag{2}$$

where $[\cdot,\cdot]$ denotes the concatenation operation along the spatial dimension, $\delta$ is a non-linear activation function. Then $f$ is split along the spatial dimension into two separate tensors $f^h$ and $f^w$, followed by another two 1×1 convolutional functions $F_h$ and $F_w$, which can be described as follows:

$$g^h = \sigma(F_h(f^h)) \ , \ g^w = \sigma(F_w(f^w)) \tag{3}$$

where $\sigma$ denotes the Sigmoid activation function. Finally, the output Y can be calculated as:

$$y_c(i,j) = x_c(i,j) \times g_c^h(i) \times g_c^w(j) \tag{4}$$

In this paper, we insert the CoordAttention module after each ShuffleLayer and further form the novel backbone ShuffleCANet. Benefiting from the attention mechanism, it is still lightweight but can pay more attention to the targets and enhance the feature expression ability in complex backgrounds.

**Bidirectional Feature Pyramid Network (BiFPN).** Feature pyramid network is designed for object detection algorithms to enable the extraction of features in different scales. It utilizes feature maps of different stages and further forms a feature pyramid network to detect objects of divergent scales.

In this paper, we use BiFPN [6] as the feature fusion neck with three optimizations based on the original PANet: Firstly, the nodes with only one input edge are removed because they have little contribution to the feature network; Secondly, an additional edge is added between the original input and output nodes at the same level to fuse more features; Thirdly, each bidirectional path is treated as one feature network layer, which is repeated multiple times to enable more high-level feature fusion. The BiFPN architecture can intensify the performance of the whole model by further refining and fusing output information, which serves as an effective connecting link between the backbone and prediction head. The structure of BiFPN in this work is shown in Fig. 6.

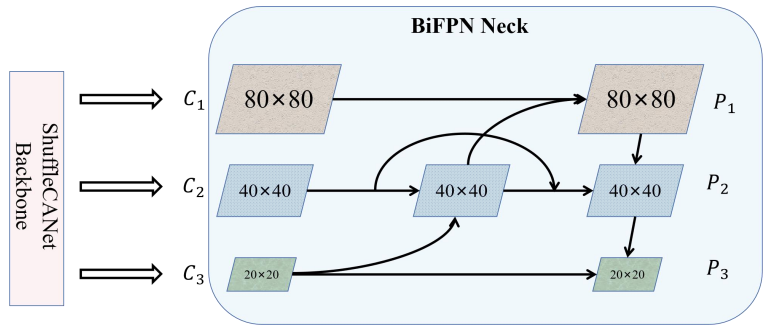

**Fig. 6.** The structure of BiFPN. $C_{1\sim3}$ and $P_{1\sim3}$ denote the input and output feature maps, respectively.

### 3.3 Loss Function

The total loss in this work includes three parts: confidence, classification, and localization loss. We use BCELoss for confidence loss and classification loss while α-CIoU loss for localization loss.

α-IoU [7] is a new family of loss generalized from the existing IoU based losses and α-CIoU is one of them generalized from CIoU. α-IoU has a power IoU term and an additional power regularization term with a single power parameter α, which significantly surpasses the existing IoU based losses and is more robust to lightweight models and noisy bounding boxes. IoU means Intersection over Union between the predicted box ($B$) and ground-truth box ($B^{gt}$), which can be defined as:

$$IoU = \frac{B \cap B^{gt}}{B \cup B^{gt}} \tag{5}$$

CIoU is an improved version of IoU that considers overlapping area, central point distance, and aspect ratio. Thus it converges faster and is more efficient in performing bounding box regression. CIoU loss can be defined as follows:

$$L_{CIoU} = 1 - IoU + |\frac{\rho^2(b, b^{gt})}{c^2}| + \beta v \tag{6}$$

where $b$ and $b^{gt}$ denote the central point of $B$ and $B^{gt}$, $\rho(\cdot)$ denotes the Euclidean distance, and $c$ is the diagonal length of the smallest enclosing box covering the two boxes. $\beta$ is the trade-off parameter and $v$ measures the consistency of the aspect ratio. $v$ and $\beta$ can be defined as follows:

$$v = \frac{4}{\pi^2}(arctan\frac{\omega^{gt}}{h^{gt}} - arctan\frac{\omega}{h})^2 \ , \ \beta = \frac{v}{1-IoU+v} \tag{7}$$

Based on the CIoU loss, α-CIoU loss is utilized in this paper as the loss function of localization to improve the bounding box regression accuracy. The function can be defined as follows:

$$L_{\alpha-CIoU} = 1 - IoU^\alpha + |\frac{\rho^{2\alpha}(b, b^{gt})}{c^{2\alpha}}| + (\beta v)^\alpha \tag{8}$$

where $\alpha$ is set as 3 in this paper.

## 4 Experiments and Results

### 4.1 Dataset

The AIZOO face mask dataset is a public dataset provided by AIZOOTech [8], which combines Wider Face [32] and Masked Faces [33] datasets with three classes: face, masked face, and mask. It comprises 7959 images in total, including 6120 images for training and 1839 images for testing. The training set consists of 13,593 faces, of which approximately 78% are exposed faces, while the test set contains 3062 faces, of which about 65% are exposed faces.

### 4.2 Evaluation Metrics

For object detection tasks, traditional evaluation metrics include precision, recall, and mean average precision (mAP); precision (P) and recall (R) can be expressed as follows:

$$P = \frac{TP}{TP+FP} \ , R = \frac{TP}{TP+FN} \tag{9}$$

where TP, FP, and FN denote true positive, false positive, and false negative, respectively. To calculate the mAP, the average precision (AP) of each class must be measured, which can be calculated by all-point interpolation approach to smooth the precision and recall curve. Furthermore, mAP can be computed by taking the mean of AP against each class. The formulas can be expressed as follows:

$$AP = \sum_{i=1}^{n}(R_{i+1} - R_i) \max_{\widetilde{R}:\widetilde{R} \geq R_{i+1}} P(\widetilde{R}) , mAP = \frac{1}{N}\sum_{i=1}^{N} AP_i \qquad (10)$$

where $i$ is the index value, P and R denote precision and recall, respectively, $N$ is the number of classes and $AP_i$ is the average precision of the $i$-th class.

In this paper, AP-Face, AP-Mask, and mAP@0.5 are used to evaluate the performance of the model, where AP-Face and AP-Mask denote APs for faces and masks, respectively. Parameter size and inference time per image are also used to evaluate the weight and speed of the model.

### 4.3 Implementation Details

The environment configuration of our work is based on Pytorch 1.9.0 framework with CUDA 11.4 and CUDNN 8.2. The models are trained on NVIDIA RTX A5000 GPU (24GB) and Xeon Gold 6139 CPU. In the training phase, we employ SGD optimizer with an initial learning rate of 1E-2 and a final learning rate of 1E-5, with the weight decay as 5E-3. Furthermore, we use three warm up epochs with a momentum of 0.8, and also apply CosineAnnealingLR method to decay the learning rate. Each experiment is trained for 200 epochs with a batch size of 16.

### 4.4 Ablation Study

To demonstrate the effectiveness of the proposed modifications based on YOLOv5, in this subsection we perform the ablation studies on ShuffleNetV2, CoordAttention, BiFPN and α-CIoU loss on the AIZOO dataset with YOLOv5s as a baseline. Precision, number of parameters, and inference time per image are used as evaluation metrics. The experiment results are summarized in Table 1.

**Table 1.** Ablation study on AIZOO dataset.

| Methods | | | Precision | Parameters | Inference time per image |
| --- | --- | --- | --- | --- | --- |
| Backbone | Neck | Loss | | | |
| BottleCSP (original) | Original | CIoU | 94.73 | 7.028MB | 1.523ms |
| ShuffleNetV2 | Original | CIoU | 94.57 | **3.274MB** | **1.003ms** |
| ShuffleNetV2 | BiFPN | CIoU | 94.82 | 3.307MB | 1.018ms |
| ShuffleCANet(+CoodAttention) | BiFPN | CIoU | 95.06 | 3.319MB | 1.095ms |
| ShuffleCANet | BiFPN | α-CIoU | **95.31** | 3.319MB | 1.092ms |

**ShuffleNetV2.** By replacing the original backbone of YOLOv5 with the modified lightweight backbone ShuffleNetV2, the size of parameters and the inference time per image are reduced by approximately 53% and 34%, respectively, while the precision is reduced by only 0.16%. The results indicate that using ShuffleNetV2 as a backbone is more likely to be applied in practice and deployed on relevant equipment.

**BiFPN.** By replacing the PANet feature fusion neck with BiFPN on the basis of ShuffleNetV2 backbone, the precision is improved by 0.35%, which surpasses the original YOLOv5. Meanwhile, it keeps the parameter size almost the same. The results indicate that BiFPN neck can better fuse features for face mask detection in this experiment.

**CoordAttention.** By integrating CoordAttention mechanism into ShuffleNetV2 to further form a novel backbone called ShuffleCANet, the precision increases by 0.24% with only a few extra parameters. Moreover, the model is still extremely fast and may be able to better extract features and focus on useful information.

**α-CIoU.** By replacing the localization loss function with α-CIoU loss in the training phase, the precision further increases by 0.25% based on ShuffleCANet and BiFPN structures. It demonstrates that α-CIoU can make bounding boxes regress better and obtain higher quality anchors.

### 4.5  Comparative Results with other Models on AIZOO

In this subsection, we compare the performance of the proposed model on AIZOO dataset with other seven state-of-the-art models used for face mask detection. The seven models include five one-stage models, namely the baseline from AIZOOTech [8], YOLOv3 [16], RetinaFace [34], RetinaFaceMask [2], and SL-FMDet [30]; and two two-stage models, namely Faster R-CNN [13] and MSAF R-CNN [35]. The comparison results are shown in Table 2.

**Table 2.** The comparative results with other SOTA methods on AIZOO dataset.

| Methods | AP-Face | AP-Mask | mAP |
|---|---|---|---|
| Baseline [8] | 89.6 | 91.9 | 90.8 |
| YOLOv3 [16] | 92.6 | 93.7 | 93.1 |
| RetinaFace [34] | 92.8 | 93.1 | 93.0 |
| RetinaFaceMask - MobileNet [2] | 93.5 | 90.4 | 92.0 |
| SL-FMDet [30] | 93.5 | 94.0 | 93.8 |
| Faster R-CNN - MobileNet [13] | 89.9 | 89.7 | 89.8 |
| MSAF R-CNN [35] | 90.3 | 90.4 | 90.4 |
| Proposed work | **94.5** | **96.2** | **95.2** |

The comparative results demonstrate that the proposed model achieves the best performance in terms of both AP-Face, AP-Mask, and mAP. Compared with the baseline model, our model increases the AP-Face, AP-Mask, and mAP by 3.9%, 4.3%, and 4.4%, respectively, which is a qualitative leap. Compared with YOLOv3 and RetinaFace, the proposed model not only achieves higher precision, but is also lighter and faster because of the light backbone. Moreover, RestinaFaceMask with MobileNet and SL-FMDet are lightweight models with a small number of parameters, but they exhibit inferior performance compared with our model. Faster R-CNN and MSAF R-CNN are two-stage models. However, their performance is not satisfactory.

The experimental results are shown in Fig. 7. Additionally, we also illustrate the performance of our model visually. As shown in Fig. 8, we select some representative images especially in public areas, the qualitative results of which demonstrate the robustness and effectiveness of our model.

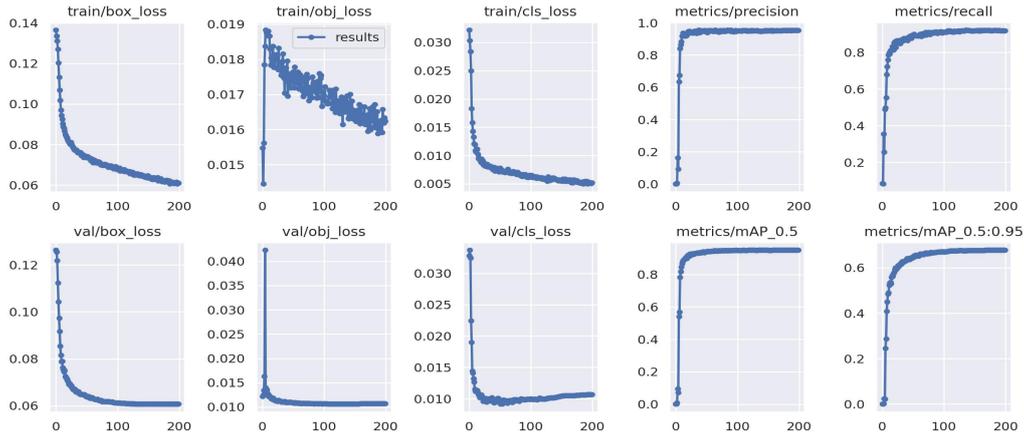

**Fig. 7.** The resulting matrix graph of the proposed work.

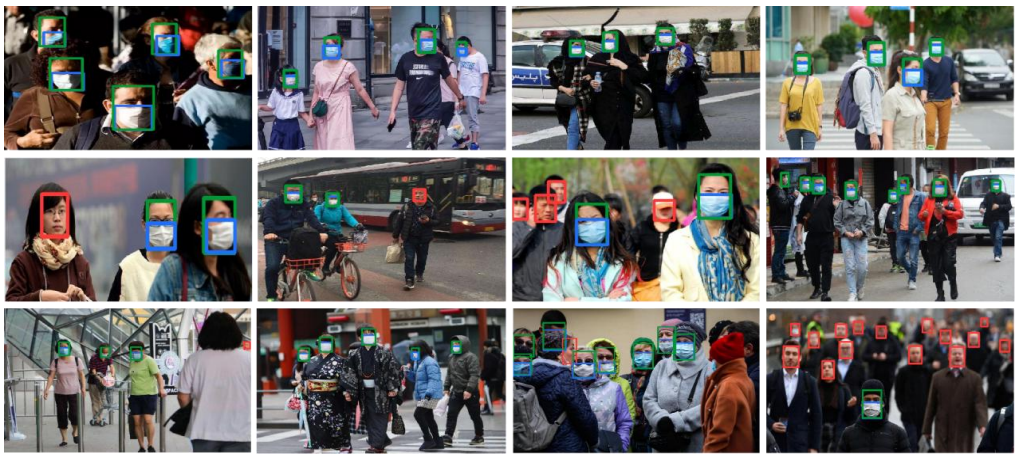

**Fig. 8.** Some detection results in real-life public situations with the proposed model. Red box denotes the unmasked face, green box denotes the masked face and blue box denotes the mask.

## 5    Conclusion

In this paper, we proposed a novel lightweight and high-performance face mask detector based on YOLOv5 with several techniques, which can achieve an excellent balance between precision and speed. To be specific, we proposed a novel backbone called ShuffleCANet that combines ShuffleNetV2 and Coordinate Attention mechanism to extract rich features and focus on useful information. After that, we used BiFPN path aggression network for feature fusion to sufficiently merge the high-level semantic information and low-level details. Additionally, we utilized α-CIoU loss to replace the original localization loss function in model training phase to obtain higher-quality anchors. The proposed work achieved state-of-the-art performance on the public face mask dataset AIZOO. Experimental results showed that our model outperformed the seven other existing face mask detectors in terms of both AP-Face, AP-Mask, and mAP. Moreover, our model is lightweight and fast, with an inference time of only 1.092ms per image on RTX A5000. Therefore, we believe that the proposed model can be employed in practice to help curb the spread of COVID-19 and thus contribute to public health.

In future work, we will handle the problem of not recognizing whether the mask is worn correctly in this work. In addition, we will focus on the detection performance of smaller faces in dense public conditions to increase its practicability.